\documentclass[12pt]{article} 
\usepackage[margin=0.75in]{geometry}
\usepackage{cite}
\newcommand{\be}{\begin{eqnarray}}
\newcommand{\ee}{\end{eqnarray}}
\usepackage[auth-sc]{authblk}
\usepackage{graphicx}
\usepackage{amsmath}
\begin{document}
\title{Can Transfer Entropy Infer Information Flow in Neuronal Circuits for Cognitive Processing?}

% Author Orchid ID: enter ID or remove command
\newcommand{\orcidauthorA}{0000-0002-2915-9504} % Add \orcidA{} behind the author's name
\author[1,3]{Ali Tehrani-Saleh}
\author[2,3,4,*]{Christoph Adami}

\affil[1]{Department of Computer Science and Engineering}
\affil[2]{Department of Microbiology \& Molecular Genetics}
\affil[3]{Department of Physics \& Astronomy}
\affil[4] {BEACON Center for the Study of Evolution\\Michigan 
State University, East Lansing, MI 48824}
\affil[*]{adami@msu.edu}

\date{}

\maketitle
\abstract{To infer information flow in any network of agents, it is important first and foremost to establish causal temporal relations between the nodes. Practical and automated methods that can infer causality are difficult to find, and the subject of ongoing research. While Shannon information only detects correlation, there are several information-theoretic notions of ``directed information'' that have successfully detected causality in some systems, in particular in the neuroscience community. However, recent work has shown that some directed information measures can sometimes inadequately estimate the extent of causal relations, or even fail to identify existing cause-effect relations between components of systems, especially if neurons contribute in a cryptographic manner to influence the effector neuron. Here, we test how often cryptographic logic emerges in an evolutionary process that generates artificial neural circuits for two fundamental cognitive tasks: motion detection and sound localization. We also test whether activity time-series recorded from behaving digital brains can infer information flow using the transfer entropy concept, when compared to a ground-truth model of causal influence constructed from connectivity and circuit logic. 
Our results suggest that transfer entropy will sometimes fail to infer causality when it exists, and sometimes suggest a causal connection when there is none. However, the extent of incorrect inference strongly depends on the cognitive task considered. These results emphasize the importance of understanding the fundamental logic processes that contribute to information flow in cognitive processing, and quantifying their relevance in any given nervous system.}

% Keywords
%\keyword{transfer entropy; information flow; causality; neural processing; motion detection; sound localization}

%\setcounter{secnumdepth}{4}
%%%%%%%%%%%%%%%%%%%%%%%%%%%%%%%%%%%%%%%%%%

%%%%%%%%%%%%%%%%%%%%%%%%%%%%%%%%%%%%%%%%%%

\section{Introduction}
When searching for common foundations of cortical computation, more and more emphasis is being placed on information-theoretic descriptions of cognitive processing~\cite{PhillipsSinger1997,RivoireLeibler2011,Adami2012,Oizumietal2014,Wibraletal2015}. One of the core tasks in the analysis of cognitive processing is to follow the flow of information within the nervous system, by finding cause-effect components. Indeed, understanding causal relationships is considered to be fundamental to all natural sciences~\cite{bunge1959causality}. However, inferring causal relationships and separating them from mere correlations is difficult, and the subject of ongoing research~\cite{granger1969investigating,pearl2000causality,pearl2009causality,sun2014causation,albantakis2019caused}. The concept of \textit{Granger causality} is an established statistical measure that aims to determine directed (causal) functional interactions among components or processes of a system. The main idea is that if a process $X$ is influencing process $Y$, then an observer can predict the future state of $Y$ more accurately given the history of both $X$ and $Y$ (written as $X_{t-k:t}$ and $Y_{t-l:t}$, where $k$ and $l$ determine how many states from the past of $X$ and $Y$ are taken into account) compared to only knowing the history of $Y$. Schreiber~\cite{schreiber2000measuring} described Granger causality in terms of information theory by introducing the concept of \textit{transfer entropy} (TE), positing that the influence of process $X$ on $Y$ can be captured using the transfer entropy from process $X$ to $Y$:
\be
\label{eq:te}
TE_{X \rightarrow Y}& =&I(Y_{t+1}:X_{t-k:t} \vert \: Y_{t-l:t})=H(Y_{t+1} \vert \: Y_{t-l:t}) - H(Y_{t+1} \vert Y_{t-l:t}, X_{t-k:t})=\nonumber\\ 
&=&\sum_{y_{t+1}} \sum_{x_{t-k:t}} \sum_{y_{t-l:t}} p(x_{t+1},x_{t-k:t},y_{t-l:t}) log {\frac{p(x_{t+1} \vert x_{t-k:t},y_{t-l:t})}{p(x_{t+1} \vert p(x_{t-k:t})} } \;.
\ee
Here as before, $X_{t-k:t}$ and $Y_{t-l:t}$ refer to the history of the states $X$ and $Y$, while $Y_{t+1}$ is the state at $t+1$ only. Further, $p(x_{t+1},x_{t-k:t},y_{t-l:t})$ is the joint probability of the state $X_{t+1}$ and the histories $X_{t-k:t}$ and $Y_{t-l:t}$, while $p(x_{t+1} \vert x_{t-k:t},y_{t-l:t})$ and $p(x_{t+1} \vert p(x_{t-k:t})$ are conditional probabilities.

The transfer entropy (\ref{eq:te}) is a conditional mutual entropy, and quantifies what the process $Y$ at time $t+1$ knows about the process $X$ up to time $t$, given the history of $Y$ up to time $t$. More colloquially, $TE_{X\rightarrow Y}$ measures ``how much uncertainty about the future course of $Y$ can be reduced by the past of $X$, given $Y$'s own past.'' Transfer entropy reduces to Granger causality for so-called ``auto-regressive processes''~\cite{Barnettetal2009}, which encompasses most biological dynamics. As a result, transfer entropy has become one of the most widely used directed information measures, especially in neuroscience (see~\cite{vicente2011transfer,wibral2014transfer,Wibraletal2015} and references cited therein). 

The use of transfer entropy to search for and detect causal relations has been shown to be inaccurate in simple case studies~\cite{PhysRevLett.116.238701,janzing2013quantifying}. For example, James et. al.~\cite{PhysRevLett.116.238701} presented scenarios in which TE may either underestimate or overestimate the flow of information from one process to another. In particular, the authors present two examples of causal processes implemented with the XOR (exclusive OR, $\oplus$) logic operation, to show that TE may underestimate or overestimate information flow from a variable to another and furthermore, it may fail to attribute the flow of information from a source variable to a receiver variable.

One of the key ideas behind the criticism by James et al. is that causal relations cannot correctly be captured in networks with cryptographic dependencies, where more than one variable causally influences another. For instance, if $Z_{t+1}=X_{t} \oplus Y_{t}$ it is not possible to determine the influence of variable $X$ on $Z$ using $TE_{X \rightarrow Z}$, which considers $X$ in isolation and independent of variable $Y$. We should make it clear that it is not the formulation of TE that is the source of these misestimations. Rather, by definition Shannon's mutual information, $I(X:Y) = H(X)+H(Y)-H(X,Y)$ is dyadic, and cannot capture polyadic correlations. Consider for example a time-independent process between binary variables $X$, $Y$, and $Z$ where $Z = X \oplus Y$. As is well-known, the mutual information between $X$ and $Z$, and also between $Y$ and $Z$ vanishes: $I(X:Z)=0, I(Y:Z)=0$ (this corresponds to the one-time pad, or Vernam cipher~\cite{shannon1949communication}). Thus, while the TE formulation aims to capture a directed (causal) dependency of variables, Shannon information measures the {\em undirected} (correlational) dependency of two variables only. As a consequence, problems with TE measurements of causality are unavoidable when using Shannon information, and do not stem from the formulation of transfer entropy~\cite{schreiber2000measuring} or similar measures such {\em causation entropy}~\cite{sun2014causation} to capture causal relations. Note that methods such as \textit{partial information decomposition} have been proposed to take into account the synergistic influence of a set of variables on the others \cite{williams2010nonnegative}. However, such higher order calculations are more costly (possibly exponentially so) and require significantly more data in order to perform accurate measurements.

\begin{figure}
\centering
\includegraphics[width=3in]{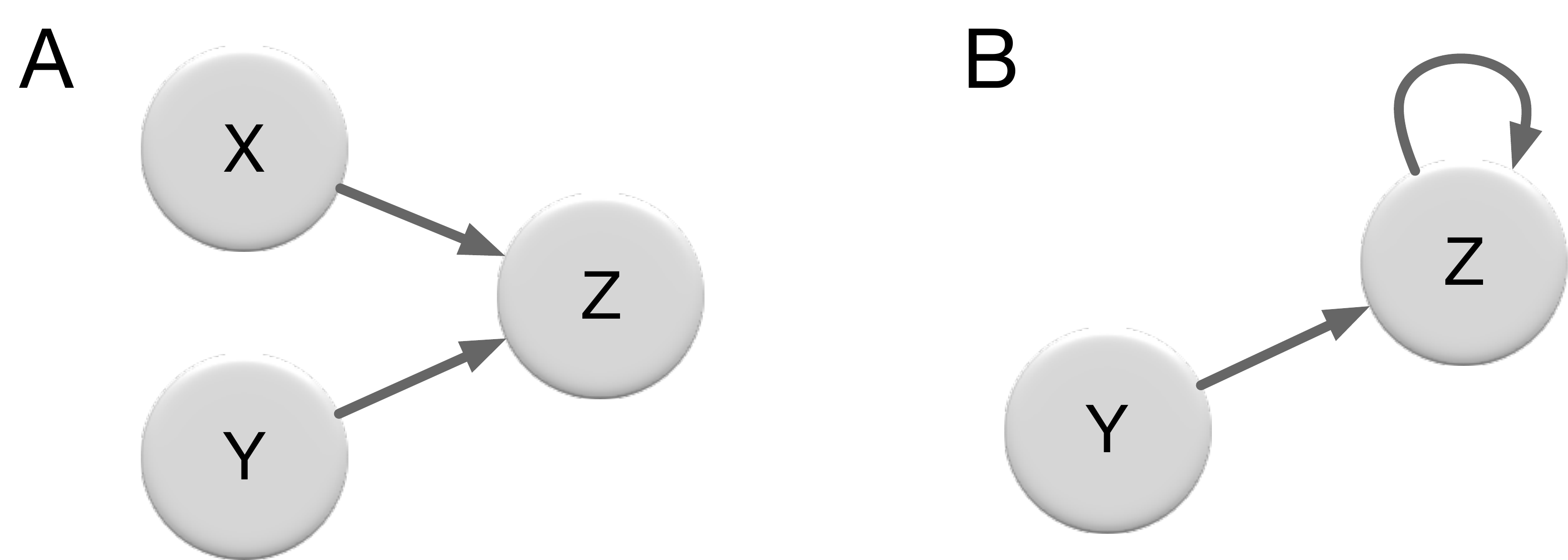}
\caption{(A) A network where processes $X$ and $Y$ influence future state of $Z$. (B) A feedback network in which processes $X$ and $Y$ influence future state $X$.}
\label{xyz-net}
\end{figure}

Before testing the performance of TE to infer information flow in cognitive networks, we can first ask how well TE captures causality in a first-order Markov process, i.e., how well TE correctly attributes the influence of inputs on the output for all 16 possible connected 2-to-1 binary relations (logic gates) with inputs $X$ and $Y$ and output $Z$ where the state of variable $Z$ is independent of its past, and inputs $X$ and $Y$ take states 0 and 1 with equal probabilities, i.e., $P(X=0)=P(X=1)=P(Y=0)=P(Y=1)=0.5$ (Fig.~\ref{xyz-net}A).

In Ref.~\cite{PhysRevLett.116.238701} James et al.\ showed that for XOR (and similarly for XNOR) relation the transfer entropies ${\rm TE}_{X \rightarrow Z}={\rm TE}_{Y \rightarrow Z}=0$,
which would imply that no information is being transferred from inputs $X$ and $Y$ to $Z$. Thus, the transfer entropies ${\rm TE}_{X \rightarrow Z}$ and ${\rm TE}_{Y \rightarrow Z}$ fail to account for the entropy of $Z$, $H(Z)=1$. Therefore, for those gates, TE does not capture information flow. 
We can extend this analysis to all the 2-to-1 gates and study when (and to what degree) information flow is incorrectly estimated using TE measurements. 
Table~\ref{table-TEs-2to1} shows the results of transfer entropy measurements for all possible 2-to-1 logic gates and the error that would occur if TE measures are used to quantify the information flow from inputs to outputs.
We find that in all other polyadic relations where both $X$ and $Y$ influence the future state of $Z$, ${\rm TE}_{X \rightarrow Z}$ and ${\rm TE}_{Y \rightarrow Z}$ capture some of the information flow from inputs to outputs, but ${\rm TE}_{X \rightarrow Z}+{\rm TE}_{Y \rightarrow Z}$ is less than the entropy of the output $Z$ by 0.19 bits (${\rm TE}_{X \rightarrow Z}+{\rm TE}_{Y \rightarrow Z}=0.62$, $H(Z)=0.81$). In the remaining 6 relations where only one of the inputs or neither of them influences the output, the transfer entropies correctly capture the causal relations. The difference between the sum of transfer entropies, ${\rm TE}_{X \rightarrow Z}+{\rm TE}_{Y \rightarrow Z}$, and the entropy of the output, $H(Z)$ in XOR and XNOR relations, stems from the fact that $I(X\!:\!Y\!:\!Z)=-1$, the tell-tale sign of encryption. Furthermore, while other polyadic gates do not implement perfect encryption, they still encrypt partially $I(X\!:\!Y\!:\!Z)=-0.19$, which we call {\em obfuscation}. It is this obfuscation that is at the heart of the TE error shown in Table~\ref{table-TEs-2to1}.

We repeated similar calculations for the case of a feedback loop network where variable $Z$ is connected to another variable Y and itself (Fig.~\ref{xyz-net}B).
These simple calculations show that in 16 relations including XOR and XNOR, the sum of the transfer entropies, ${\rm TE}_{Y \rightarrow Z}+I(Z_{t+1}:Z_t)$ (the formulation for transfer entropy of a variable to itself reduces to processed information $I(Z_{t+1}:Z_t)$) is equal to the entropy of the output $Z$. However, in XOR and XNOR relations transfer entropy incorrectly attributes all the information to one of the input variables and no influence is attributed to the other. Furthermore, in the polyadic relations other than XOR and XNOR, the transfer entropies ${\rm TE}_{Y \rightarrow Z}$ and $I(Z_{t+1}:Z_t)$ differ in value while variables $X$ and $Y$ equally influence the state of the output $Z$. James et. al. also have argued that it may be inaccurate to quantify the information flow from $Y$ to $Z$ with ${\rm TE}_{Y \rightarrow Z}$.

%Fig. 1

\begin{table}
\caption{Transfer entropies and information in all possible 2-to-1 binary logic gates with or without feedback. The logic of the gate is determined by the value $Z_{t+1}$ (second column) as a function of the input $X_t Y_t$=(00,01,10,11). $H(Z_{t+1})$ is the Shannon entropy of the output assuming equal probability inputs, $TE_{X \rightarrow Z}$ is the transfer entropy from $X$ to $Z$. In 2-to-1 gates without feedback, transfer entropies ${\rm TE}_{X \rightarrow Z}$ and ${\rm TE}_{Y \rightarrow Z}$ reduce to $I(X_t\!:\!Z_{t+1})$, and $I(X_t\!:\!Z_{t+1})$, respectively. Similarly, transfer entropy of a process to itself is simply $I(Z_t\!:\!Z_{t+1})$ which is the information processed by $Z$.}
\label{table-TEs-2to1}
\begin{tabular}{|c|c|c||c|c|c||c|c|c|}
%\toprule
\cline{4-9}
\multicolumn{3}{c||}{} & \multicolumn{3}{c||}{2-to-1 network, $Z=f(X,Y)$} & \multicolumn{3}{c|}{2-to-1 feedback loop, $Z=f(Y,Z)$} \\ \hline
gate & $Z_{t+1}$ & $H(Z_{t+1}$) & ${\rm TE}_{X \rightarrow Z}$ & ${\rm TE}_{Y \rightarrow Z}$ & TE error & ${\rm TE}_{Y \rightarrow Z}$) & $I(Z_t:Z_{t+1})$ & TE error \\ \hline 
%\midrule
ZERO    & (0,0,0,0)  & 0.0  & 0.0  & 0.0  & 0.0  & 0.0  & 0.0  & 0.0    \\
AND     & (0,0,0,1)  & 0.81 & 0.31 & 0.31 & 0.19 & 0.5 & 0.31 & 0.19   \\
AND-NOT & (0,0,1,0)  & 0.81 & 0.31 & 0.31 & 0.19 & 0.5 & 0.31 & 0.19   \\
AND-NOT & (0,1,0,0)  & 0.81 & 0.31 & 0.31 & 0.19 & 0.5 & 0.31 & 0.19   \\
NOR     & (1,0,0,0)  & 0.81 & 0.31 & 0.31 & 0.19 & 0.5 & 0.31 & 0.19   \\
COPY    & (0,0,1,1)  & 1.0  & 1.0  & 0.0  & 0.0  & 1.0  & 0.0  & 0.0    \\
COPY    & (0,1,0,1)  & 1.0  & 0.0  & 1.0  & 0.0  & 0.0  & 1.0  & 0.0    \\
XOR     & (0,1,1,0)  & 1.0  & 0.0  & 0.0  & 1.0  & 1.0  & 0.0  & 1.0    \\
XNOR    & (1,0,0,1)  & 1.0  & 0.0  & 0.0  & 1.0  & 1.0  & 0.0  & 1.0    \\
NOT     & (1,0,1,0)  & 1.0  & 0.0  & 1.0  & 0.0  & 0.0  & 1.0  & 0.0    \\
NOT     & (1,1,0,0)  & 1.0  & 1.0  & 0.0  & 0.0  & 1.0  & 0.0  & 0.0    \\
OR      & (0,1,1,1)  & 0.81 & 0.31 & 0.31 & 0.19 & 0.5 & 0.31 & 0.19   \\
OR-NOT  & (1,0,1,1)  & 0.81 & 0.31 & 0.31 & 0.19 & 0.5 & 0.31 & 0.19   \\
OR-NOT  & (1,1,0,1)  & 0.81 & 0.31 & 0.31 & 0.19 & 0.5 & 0.31 & 0.19   \\
NAND    & (1,1,1,0)  & 0.81 & 0.31 & 0.31 & 0.19 & 0.5 & 0.31 & 0.19   \\
ONE     & (1,1,1,1)  & 0.0  & 0.0  & 0.0  & 0.0  & 0.0  & 0.0  & 0.0    \\
\hline
\end{tabular}
\end{table}

Given that TE measurements only fail to correctly identify causal relations in cryptographic gates and demonstrate partial errors in polyadic relations, we now set out to determine how often these relations appear in basic cognitive tasks, and how much error do they introduce in transfer entropy measurements. If the total error in transfer entropy measurements of information flow in cognitive networks is significant, an analysis of causal relationships among neural components (neurons, voxels, etc.) using this concept is bound to be problematical. If, however, these errors are reasonably low within biological control structures because cryptographic logic is rarely used, then treatments using the TE concept can largely be trusted.

To answer this question, we use a new tool in computational cognitive neuroscience, namely computational models of cognitive processing that can explain task-performance in terms of plausible dynamic components~\cite{KriegeskorteDouglas2018}. In particular, we use Darwinian evolution to evolve artificial digital brains (also known as Markov Brains or MBs~\cite{hintze2017markov}) that can receive sensory stimuli from the environment, process this information, and take actions in response\footnote{In the following we refer to digital brains as ``Brains", while biological brains remain ``brains".}. We evolve Markov Brains that perform two different cognitive tasks whose circuitry is thoroughly studied: visual motion detection~\cite{borst1989principles}, as well as sound localization~\cite{moore2012introduction,pickles2013introduction}. Markov Brains have been shown to be a powerful platform that can unravel the information-theoretic correlates of fitness and network structure in neural networks~\cite{edlund2011integrated,Albantakisetal2014,schossau2015information,marstaller2013evolution,Jueletal2019}. This computational platform enables us to analyze structure, function, and circuitry of hundreds of evolved digital Brains. As a result, we can obtain statistics on the frequency of different causal relations in evolved circuits (as opposed to studying only a single evolutionary outcome), and further assess how crucial different operators are for each evolved task, by performing knockout experiments in order to measure an operator's contribution to the task. In particular, we first investigate the composition of different types of logic gates in networks evolved for the two cognitive tasks, and then theoretically estimate how accurate transfer entropy measures could be when applied to quantify the information flow from a process to another in such simple cognitive networks. We then use transfer entropy measures as a proxy to identify causal relations between neurons of evolved circuits using the time series of neural recordings obtained from behaving brains engaged in their task, and evaluate how successful transfer entropy is in detecting causal relations.
While artificial evolution of control structures (``artificial Brains'') is not a substitute for the analysis of information flow in biological brains, this investigation should provide some insights on how accurate (or inaccurate) transfer entropy measures could be

\section{Materials and Methods}

\subsection{Markov Brains}
Markov Brains (MB) are evolvable networks of binary neurons (they take value 0 for a quiescent neuron, or 1 for firing neuron) in which neurons are connected via probabilistic or deterministic logic gates (in this work, we constrain MBs to only use 2-to-1 deterministic logic gates). The states of the neurons are updated in a Markov process, i.e., the probability distribution of states of the neurons at time step $t+1$ depends only on the states of neurons at time step $t$. This does not imply that Markov Brains are memoryless, because the state of one neuron can be stored by repeatedly writing into its own (or another) neuron's state variable~\cite{edlund2011integrated,marstaller2013evolution,hintze2017markov}. The connectivity and the underlying logic of the MB's neuronal network is encoded in a genome. Thus, we can evolve populations of MBs using a Genetic Algorithm (GA)~\cite{Michalewicz1996} to perform a variety of cognitive tasks (for a more detailed description of Markov Brain function and implementation see~\cite{hintze2017markov}). In the following sections, we describe two fitness functions designed to evolve motion detection and sound localization circuits in MBs.

\subsection{Motion Detection}
\begin{figure}[htb]
\centering
\includegraphics[width=4.5in]{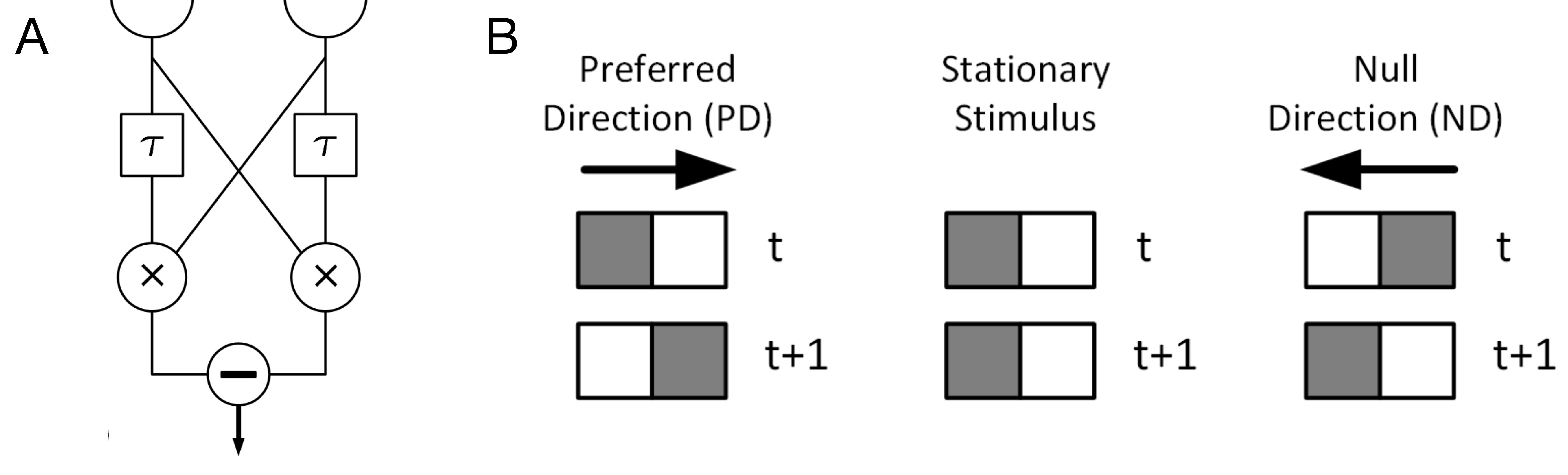}
\caption{(A) A Reichardt detector circuit. In these circuits, the results of the multiplications from each half circuit are subtracted to generate the response. (B) Schematic examples of three types of input patterns received by the two sensory neurons at two consecutive time steps. Grey squares show presence of the stimuli in those neurons.}
\label{md-experiment}
\end{figure}

The first fitness function is designed in order to evolve MBs that function as a visual motion detection circuit. Reichardt and Hassenstein proposed a circuit model of motion detection that is based on a delay-and-compare scheme~\cite{hassenstein1956systemtheoretische}. The main idea behind this model is that a moving object is sensed by two adjacent receptors on the retina, at two different time points. Fig.~\ref{md-experiment} shows the schematic of a Reichardt detector in which the $\tau$ components delay the stimulus and $\times$ components multiply the signals, i.e., fires if the signal from the receptor and delay component arrive at the same time. The result of the multiplication units for two different directions is then subtracted so that high values denote motion in one direction (the ``preferred direction", PD), low values denote the opposite direction (null direction, ND), and intermediate values encode a stationary stimulus.
The experimental setup for evolution of motions detection circuits is similar to the setup previously used in~\cite{Tehranietal2018}. In that setup, two sets of inputs are presented to a MB at two consecutive times and the Brain classifies the input as preferred direction (PD), stationary, or null direction (ND). The value of the sensory neuron becomes 1 when a stimulus is present, and it becomes 0 otherwise (see Fig.~\ref{md-experiment}). Thus, 16 possible sensory patterns can be presented to the MB to classify, among which 3 input patterns are PD, 3 are ND, and the other 10 are stationary patterns. Two neurons are assigned as output neurons of the motion detection circuit. The sum of binary values of these neurons represents the output of the motion detection circuit, 0: ND, 1: stationary stimulus, 2: PD.

\subsection{Sound Localization}
\begin{figure}[htb]
\centering
\includegraphics[width=4.5in]{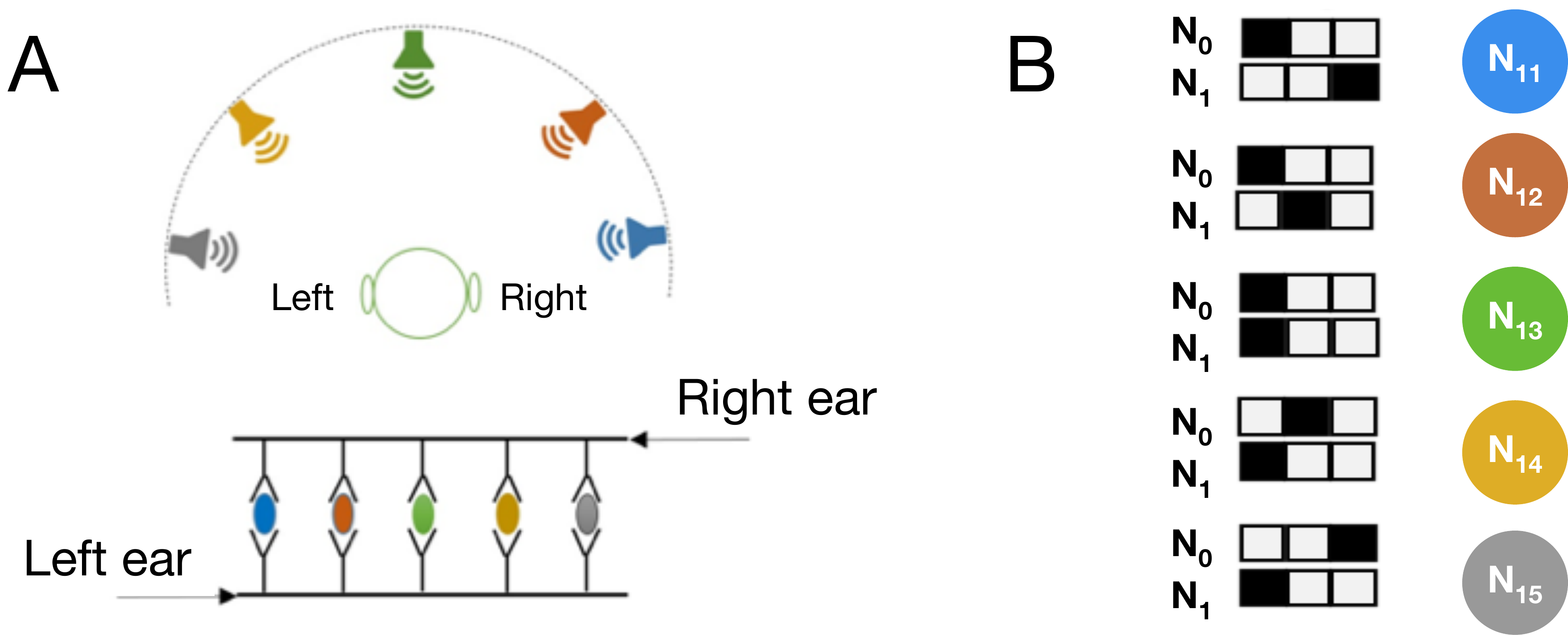}
\caption{(A) Schematic of 5 sound sources at different angles with respect to a listener (top view) and Jeffress model of sound localization. (B) Schematic examples of 5 time sequences of input patterns received by the two sensory neurons (receptors of two ears) at three consecutive time steps. Black squares show presence of the stimuli in those neurons.}
\label{SL-experiment}
\end{figure}

The second fitness function is designed to evolve MBs that function as a sound localization circuit. Sound localization mechanisms in mammalian auditory systems function based on several cues such as interaural time difference, interaural level difference, etc.~\cite{middlebrooks1991sound}. Interaural time difference (which is the main cue behind the sound localization mechanism) is the difference between the times sound reaches two ears. Fig.~\ref{SL-experiment}A shows a simple schematic of a sound localization model proposed by Jeffress~\cite{jeffress1948place} in which sound reaches the right ear and left ear at two possibly different times. These stimuli are then delayed in an array of \textit{delay} components and travel to an array of detector neurons (marked with different colors in Fig.~\ref{SL-experiment}A). Each detector only fires if the two signals from different pathways, the left ear pathway (shown in bottom) and the right ear pathway (shown in top), reach that neuron simultaneously.
%Fig3

In our experimental setup, two sequences of stimuli are presented to two different sensory neurons (neurons $N_0$ and $N_1$) that represent receptors in two ears. The stimulus in two sequences are lagged or advanced with respect to one another (as shown in Fig.~\ref{SL-experiment}B). The agent receives these sequences and should identify 5 different angles from where that sound is coming from. The binary value of the sensory neuron becomes 1 when a stimulus is present, shown as black blocks in Fig.~\ref{SL-experiment}B, and it becomes 0 otherwise, shown as white blocks in Fig.~\ref{SL-experiment}B. Similar to schema shown in Fig.~\ref{SL-experiment}A, Markov Brains have five designated output neurons ($N_{11}$-$N_{15}$) and each neuron corresponds to one of the sound sources placed at a specific angle. Colors of detector neurons ($N_{11}$-$N_{15}$) in Fig.~\ref{SL-experiment}B match the angle of each sound source in Fig.~\ref{SL-experiment}A.
%%%%%%%%%%%%%%%%%%%%%%%%%%%%%%%%%%%%%%%%%%
\section{Results}
For the motion detection (MD) and sound localization (SL) tasks, we evolved 100 populations each for 10,000 generations, allowing all possible 2-to-1 (deterministic) logic gates as primitives. At the end of each evolutionary run, we isolated one of the genotypes with the highest score from each population to generate a representative circuit.
\subsection{Gate Composition of Evolved Circuits}
\begin{figure}[htb]
\centering
\includegraphics[width=0.9\columnwidth]{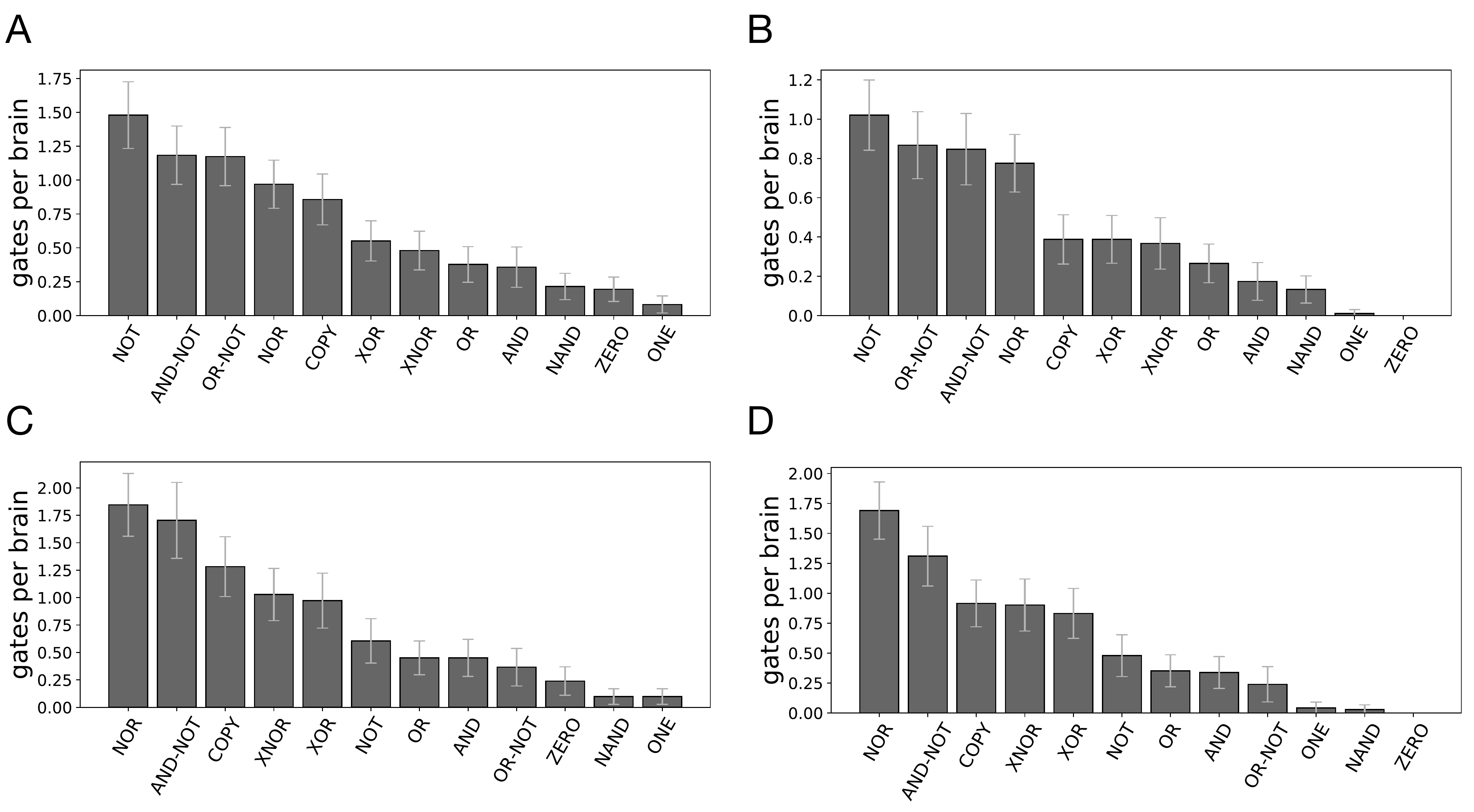}
\caption{Frequency distribution of all, as well as essential, gates in evolved Markov Brains that perform the motion detection or sound localization task perfectly. (A) All gates, motion detection. (B) Essential gates, motion detection. (C) All gates, sound localization. (D) Essential gates, sound localization.}
\label{gates-dist}
\end{figure}

Out of 100 populations evolved in motion detection task, 98 led to circuits that perform motion detection with perfect fitness. The number of gates in evolved Brains varies tremendously, with a minimum of 4 and maximum of 17 (mean=7.92, SD=2.48). The frequency distribution of types of logic gates per each individual Brain is shown for these 98 perfect circuits in Fig.~\ref{gates-dist}A (in this figure, AND-NOT is an asymmetric AND operation where one of the variables is negated, for example $X'\cdot Y$. Similarly, OR-NOT is an asymmetric OR operation, e.g. $X+Y'$). 
To gain a better understanding of the distribution of logic gates and how they compose the evolved motion detection circuits, we performed gate-knockout assays on all 98 Brains. We sequentially eliminated each logic gate and re-measured the mutant Brain's fitness, thus allowing us to estimate which gates were essential to the motion detection function (if there is a drop in mutant Brain's fitness) and which gates were redundant to the motion detection function (if a mutant Brain's fitness remains perfect). The frequency distribution of each type of logic gate per individual Brain for essential gates is shown for the 98 perfect Brains in Fig.~\ref{gates-dist}B. 

For the sound localization task, 71 evolution experiments out of 100 resulted in Markov Brains with perfect fitness. The minimum number of gates was 6, with a maximum of 15 (mean=9.14, SD=1.77). Fig.~\ref{gates-dist}C shows the frequency distribution of types of logic gates per Brain for these 71 perfect Brains.
We also performed a knockout analysis on evolved sound localization circuits on all 71 Brains. The frequency distribution of each type of logic gate per individual Brain for essential gates is shown for the 71 perfect Brains in Fig.~\ref{gates-dist}D. These results demonstrate that the gate type compositions and circuit structures in evolved Brains for motion detection (MD) and sound localization (SL) tasks are significantly different. The total number of logic gates (ignoring duplicates) in the SL task (9.14 gates per Brain, SD=1.77) is greater than the total number of gates in the MD task (7.92 gates per Brain, SD=2.48). Moreover, the number of essential gates in SL (7.13 gates per Brain, SD=1.24) is also greater than the number of essential gates in MD (5.23 gates per Brain, SD=1.31). 

%Fig. 4

\subsection{Transfer Entropy Misestimates Lower Bound on Information Flow}
As discussed before, transfer entropy measures may misestimate the information flow from input to output and may fail to correctly identify the source of information. Table~\ref{table-TEs-2to1} gave a detailed analysis of transfer entropy measurements and their misestimates for all possible 2-to-1 logic gates. Given the gate distributions of evolved circuits for motion detection and sound localization tasks, we can calculate the error that would occur when using transfer entropy to quantify the information flow from source neurons (i.e., input neurons to gates), to receiver neurons (i.e., output neurons from gates). We can also calculate what fraction of the information flow from inputs to outputs is {\em correctly} quantified by the transfer entropy in the evolved circuits. 

\begin{figure}[htb]
\centering
\includegraphics[width=0.85\columnwidth]{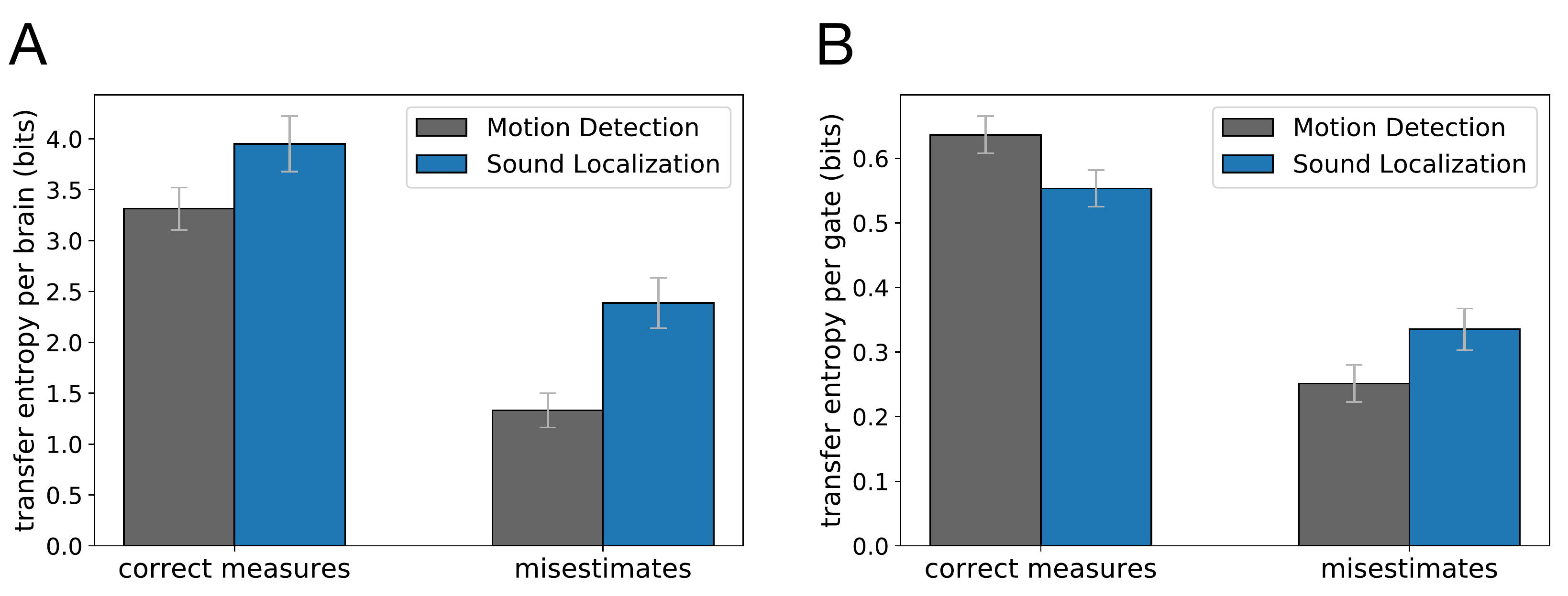}
\caption{Transfer entropy measures, exact measures and misestimates by transfer entropy, on essential gates of perfect circuits for motion detection, and sound localization task. Columns show mean values and 95\% confidence interval of misestimates and exact measures (A) per Brain, and (B) per gate.}
\label{TEs-md-sl}
\end{figure}

In our analysis, we only evaluated the contribution of gates deemed essential via the knockout test. The mean values of calculated misestimates of information flow as well as correct measurements with their 95\% confidence intervals for 98 evolved circuits that perform motion detection task, and for 71 evolved sound localization Brains are shown in Fig.~\ref{TEs-md-sl}A. In Fig.~\ref{TEs-md-sl}B, we normalized misestimates and correct measurements by dividing by the number of essential gates in each Brain, and averaged them across Brains.
It is worth noting that the calculated information flow misestimates shown in these plots are lower bounds of the error, since they are only based on the network structure and the gate composition of each Brain as well as the analytical results presented in table~\ref{table-TEs-2to1}, and do not take into account the errors that could occur as a result of factors such as sampling errors in the dataset or structural complexities in the network, such as recurrent or transitive relations~\cite{sun2014causation,albantakis2019caused}. Along the same line of reasoning, calculated values of correct measurements represent an upper bound of correct information flows that could be measured by transfer entropy.

These results further reveal that the circuit structures and gate type compositions in the two tasks are significantly different, and that this structural difference leads to different outcomes when transfer entropy measures are used to detect information flows. Transfer entropy can potentially capture 3.31 bits (SE=0.10) of information flow correctly in evolved motion detection circuits (0.64 bits per gate, SE=0.014), and 3.95 (averaged across 71 Brains, SE=0.14) bits in evolved sound localization circuits (0.55 bits per gate, SE=0.014). However, the upper bound of information flow error when using transfer entropy in evolved sound localization circuits is 2.39 bits (averaged across 71 Brains, SE=0.12) which is significantly higher than the upper bound of misestimates in evolved motion detection circuits is 1.33 bit (average across 98 Brains, SE=0.085). The upper bounds on the error is 0.25 bits (SE=0.014) in evolved motion detection circuits whereas it is 0.34 bits (SE=0.016) per gate in evolved sound localization circuits. These findings show that the accuracy of transfer entropy measurements for detecting information flow in digital neural networks can vary significantly from one task to another.

%Fig. 5

\subsection{Transfer Entropy Measurements From Recordings of Evolved Brains}
\begin{figure}[htb]
\centering
\includegraphics[width=0.90\columnwidth]{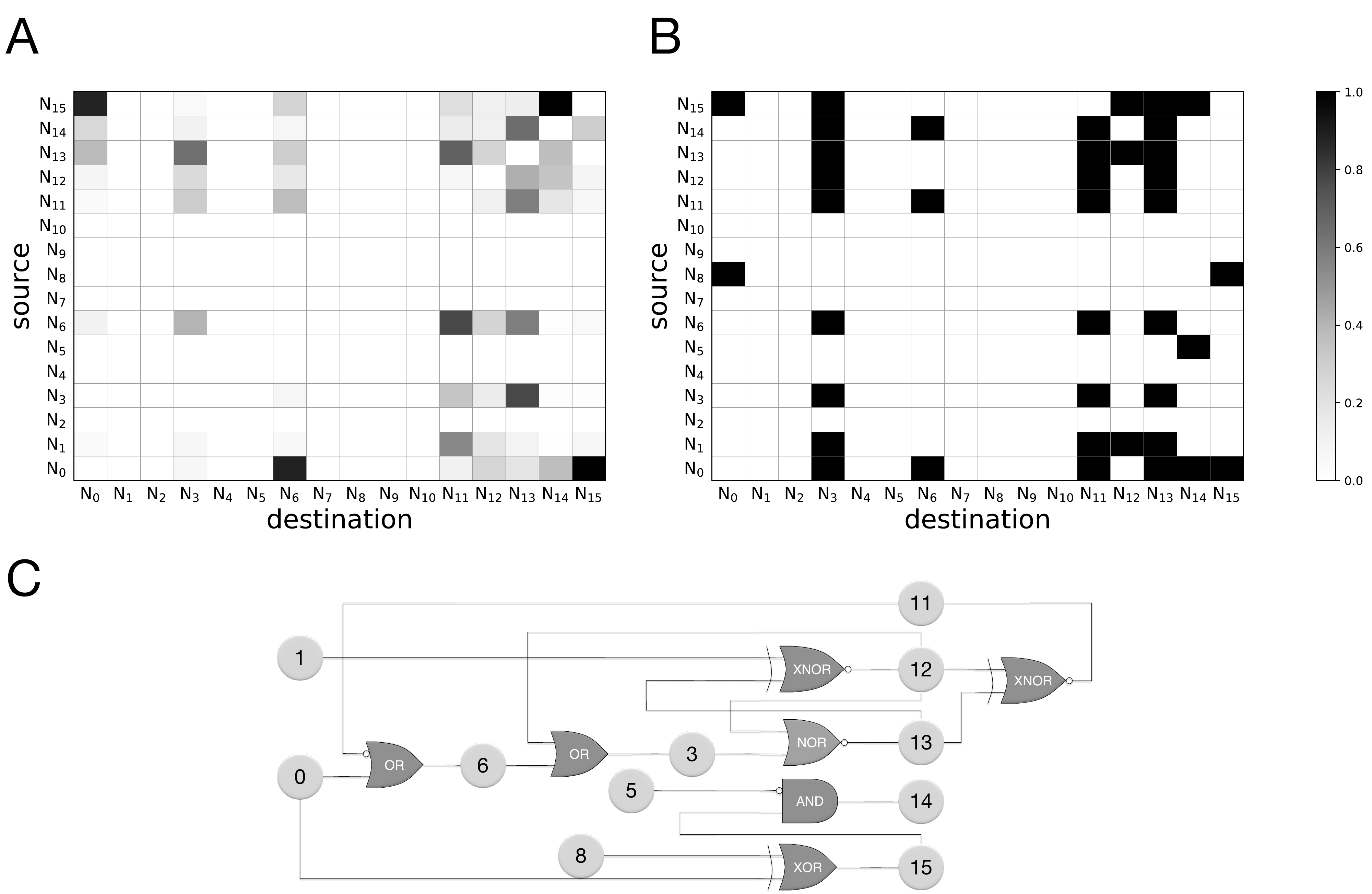}
\caption{(A) Transfer entropy measures from neural recordings of a Markov Brain evolved for sound localization. (B) Influence map (also receptive field) of neurons derived from a combination of the logic gates connections and the Boolean logic functions for the same evolved Markov Brain, shown in (C). (C) The logic circuit of the same evolved Markov Brain; neurons $N_0$ and $N_1$ are sensory neurons, and neurons $N_{11}-N_{15}$ are actuator (or decision) neurons.}
\label{exmaple-TE}
\end{figure}

In the previous section we calculated the theoretical upper bound of correct information flow measurements as well as the lower bound of misestimates when using transfer entropy, given the gate distribution for each cognitive task. Here we use a different approach to assess transfer entropy measurement accuracy in identifying inter-neuronal relations of evolved Markov Brains: we record the neural activities of an evolved Brain when performing a particular cognitive task, similar to the neural recording (``brain mapping'') performed on behaving animals. We collect the recordings in all possible trials for each cognitive task and create a dataset for each evolved Brain for that cognitive task. We then use these recordings to measure transfer entropy for every pair of neurons ${\rm TE_{N_i \rightarrow N_j}}$ in the network. These transfer entropy measures can be used as a statistic to test whether a neuron $N_i$ causally influences another neuron $N_j$. Fig.~\ref{exmaple-TE}A shows the result of TE calculations performed on neural recording for a Markov Brain evolved in the sound localization task.

To test the accuracy of the TE prediction, we construct an influence map for each neuron of the Markov Brain that shows which other neurons are influenced by a particular neuron. Such a mapping also determines the receptive field of each neuron, which specifies which other neurons influence a particular neuron. Markov Brains evolve complex networks in which multiple logic gates can write to the same neuron and as a result, it is not straightforward to deduce input-output relation among neurons. Indeed, it was previously argued that even with a complete knowledge of a given system, finding the causal relation among the components of the system may be a very difficult task~\cite{pearl2000causality,paul2013causation,halpern2016actual}. 

To create our ``ground truth" model of causal relations, we take into consideration two different components of a Brain's network. First, we take into account the input neurons of a gate and its output neuron, while we also taking into consideration the type of the logic gate. For example, in the case of a 'ZERO' gate where the output is always 0 we do not interpret this connection to have a causal influence or reflect information flow. Second, we analytically extract the binary state of each neuron as a Boolean function of all other neurons using a logic table of the entire Brain (logic table of size $2^{16}$, for 16 neurons). This helps us rule out neurons that are connected as inputs to a logic gate while not actually contributing to the output neuron of that gate. Note that this procedure is specifically helpful in cases where more than one logic gate writes into a neuron. Fig.~\ref{exmaple-TE}B shows an example of ``ground truth" influence map of neurons for a Brain evolved for sound localization. Each row of this plot shows the influence map of the corresponding neuron and each column represents the receptive field of that neuron. Note that in this plot values are binary, i.e., they are either 0 or 1 which specifies whether a source neuron influences a destination neuron, whereas TE measurements vary in the range [0, 1] bits. Keep in mind that this influence map is only an estimate of information flow gathered from gate logic and connectivity shown in Fig.~\ref{exmaple-TE}C.

%Fig. 6

\begin{figure}[htb]
\centering
\includegraphics[width=0.75\columnwidth]{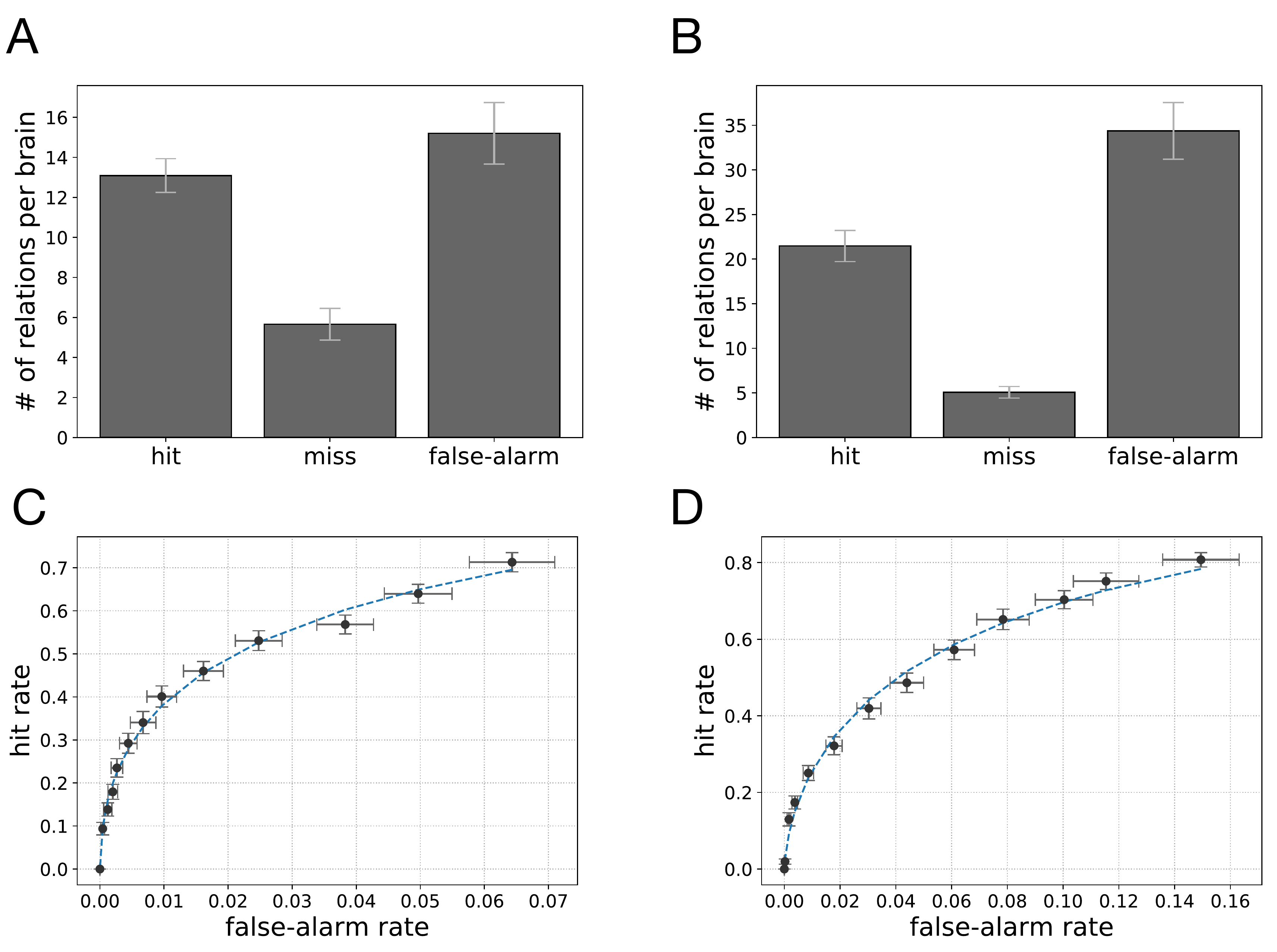}
\caption{Transfer entropy performance in detecting relations among neurons of evolved (A) motion detection circuits, (B) sound localization circuits. Presented values are averaged across best performing Brains along with 95\% confidence intervals. Receiver operating characteristic (ROC) curve representing TE performance with different thresholds to detect neurons relations in evolved (C) motion detection, (D) sound localization circuits.}
\label{TE-ROC}
\end{figure}

In order to compare TE measurements with influence maps, we first assume that any non-zero value of the ${\rm TE}_{N_i \rightarrow N_j}$ implies that there is some flow of information from neuron $N_i$ to $N_j$, and therefore, that the state of $N_i$ causally influences the state of $N_j$. We then evaluate how well TE measurements detect the causal relations of neurons based on this assumption. In particular, for each evolved Brain we count 1) the number of causal relations between neurons correctly detected by TE (hit), 2) the number of relations that are present in the influence map but were not detected by TE (miss), and 3) the number of relations detected by TE measurements as causal that according to the influence map were incorrectly detected (false-alarm). Fig.~\ref{TE-ROC}A and B show the performance results of TE measurements in detecting causal relations in Brains evolved in motion detection and sound localization, respectively (averaged across best performing Brains and 95\% confidence interval). % describe what these results show and also describe numbers.
We observe that the number of false-alarms in motion detection (mean=15.2, SE=0.77) is comparable to the number of hits (mean=13.1, SE=0.42), whereas in sound localization the number of false-alarms (mean=34.4, SE=1.59) which is significantly higher than the number of hits (mean=21.5, SE=0.88). This again underscores that the accuracy of transfer entropy measures strongly depends on the characteristics of the task that is being solved.

In the results shown in~\ref{TE-ROC}A, B, we assumed that any value of transfer entropy greater than 0 implies information flow. This assumption can be relaxed such that only transfer entropy values that are greater than a particular threshold imply information flow or causal relation. We calculated TE measurement performance for a variety of threshold values in the range [0, 1]. The results are shown as ROC (Receiver Operating Characteristic) curves (hit rates as a function of false-alarm rates as well as their 95\% confidence intervals) in Figs.~\ref{TE-ROC}C and D for motion detection and sound localization tasks, respectively~\cite{macmillan2004detection}. In these plots, the dashed line shows a fitted ROC curve assuming a Gaussian distribution for the $p(TE \vert IF={\rm present})$ and $p(TE \vert IF={\rm absent})$ ($IF={\rm present}$ denoted causal relations or existing information flow and $IF={\rm absent}$ non-existing information flow), where the resulting ROC function is $f(x)=\frac{1}{2}erfc(\frac{\mu_1-\mu_2}{\sqrt{2}\sigma_2}+\frac{\sigma_1}{\sigma_2}erfc^{-1}(2x))$ with $erfc$ the ``error function'' complement and $erfc^{-1}$ the inverse of the error function complement.

In the ROC plots, the datapoint with the highest hit rate (right-most data point) is the normalized result shown in Fig.~\ref{TE-ROC}A, B, that is, the analysis with a vanishing threshold. Note also that the data in Fig.~\ref{TE-ROC} represent hit rates against false-alarm rate for thresholds spanning the entire range [0,1], implying that hit rates cannot be increased any further unless all relations are accepted as causal relations (hit rate=false-alarm rate=1).
The false-alarm rates in the ROC curves are actually fairly low in spite of the significant number of false alarms we see in Fig.~\ref{TE-ROC}A, B. This is due to the fact that the number of existing causal relations in a Brain network is much smaller than the number of non-existing relations between any pair of neurons. Thus, when dividing the number of false-alarms by the total number of non-existing causal relations, the false-alarm rate is low.

%Fig. 7

\section{Discussion}
We used an agent-based evolutionary platform to quantitatively evaluate the accuracy of transfer entropy measurements as a proxy for measuring information flow, when applied to digital Brains. To this end, we measured the frequency and significance of cryptographic and polyadic 2-to-1 logic gates in evolved digital Brains that perform two fundamental and well-studied cognitive tasks: visual motion detection and sound localization. We evolved 100 populations for each of the cognitive tasks and analyzed the Brain with highest fitness at the end of each run. Markov Brains evolved a variety of neural architectures that vary in number of neurons and the number of logic gates, as well as the type of logic gates to perform each of the cognitive tasks. In fact, both modeling~\cite{prinz2004similar} and empirical~\cite{goaillard2009functional} studies have shown that a wide variety of internal parameters in neural circuits can result in the same functionality~\cite{marder2011variability}. Thus, it would be informative and perhaps necessary to examine a variety of circuits that perform the same cognitive task~\cite{Tehranietal2018}. %We measured the theoretical lower bound of misestimates that result from using transfer entropy measures in quantifying information flow among processes of the evolved network. These results showed that the extent of transfer entropy misestimates may be very dependent on the type of cognitive circuit to which it is applied. We also performed transfer entropy measurements on neural recording of the Markov Brains during their task and used transfer entropies as a statistic to detect causal relations among the neurons. Since the entire structure and function of these evolved circuits are accessible, we can evaluate how successful transfer entropy measures are in identifying causal relations and information flow in an evolved Brain.

In order to assess the significance of cryptographic and polyadic gates that result in misestimates of information flow in circuits, we performed gate knockout assays on Brains. And since we know exactly which neurons are influencing output neurons and the dependency between them for any particular gate, we can calculate the theoretical lower bound of misestimates by transfer entropy measures in our artificial Brains. 
The transfer entropy misestimates lower bound was 1.33 bits (SE=0.08) per Brain on average for Brains evolved in motion detection task whereas in evolved Brains performing task localization task, the misestimate lower bound was significantly higher, 2.39 bits (SE=0.12) per Brain on average.
These results suggest that evolving circuits that perform various types of cognitive tasks require different set of relations (logic gates). More importantly, these inherent differences between the two tasks result in different levels of accuracy when using transfer entropy measures to identify causal relations among neurons, and thus to follow the flow of information in these Brains. Furthermore, it is important to note that in calculating these misestimate lower bounds, we only accounted for the misestimates that result from TE measurements in polyadic or cryptographic gates. However, we commonly face several other challenges when searching for causal relations among components of nervous systems (neurons, voxels, etc.). These challenges range from intrinsic noise in neurons to inaccessibility of recording data for larger populations of neurons which we discuss in more detail later.
%emphasize they are just lower bounds and more can go wrong with noise, sampling error, not knowing optimal history length, we record from all neurons not a subset of neurons

We also tested how well transfer entropy can identify the existence of information flow between any pair of neurons using the statistics of neural recordings at two subsequent time points only. Because a perfect model for the ``ground truth'' of information flow is difficult (if not impossible) to establish, we use an approximate ground truth that uses the connectivity of the network, along with information from the (simplified) logic function to provide a comparison. We find that TE captures many of the connections established by the ground truth model, with a true positive rate (hit rate) of 69.8\% for motion detection and 80.9\% for sound localization (assuming any non-zero value of transfer entropy implies causal relation). The TE measurements miss some relations from the established ground truth while also providing demonstrably false positives, with false-alarm rate of 6.4\% in motion detection and 15.0\% for sound localization. For example, some of the information flow estimates in Fig.~\ref{exmaple-TE} manifestly reverse the actual information flow, suggesting a backwards flow that is causally impossible. Such erroneous backwards influence is possible, for example, when the signal has a periodicity that creates accidental correlations with significant frequency. Besides these false positives, the false negatives (missed inferences) are due to the use of information-hiding (cryptographic or obfuscating) relations, as discussed earlier.

It is noteworthy that in the transfer entropy measurements we performed, we benefited from multiple factors that are commonly great challenges in TE analysis of biological neural recordings. First, our TE measurement results were obtained using recordings of perfect (noise-free) neurons, while biological neurons are intrinsically noisy. We were also able to use the recordings from every neuron in the network, which presumably results in more accurate estimates. In contrast, in biological networks we only have the capacity to record from a finite number of neurons which, in turn, constrains our understanding of causal relations in the network. 
In the results presented in this work, we benefited from another factor that also made our measurements more accurate compared to TE measurements on actual recordings, namely the fact we know that our evolved Brains are 1$^{st}$-order Markov processes. 
Recall that in equation~\ref{eq:te}, transfer entropy ${\rm TE}_{X \rightarrow Y}$ is the shared information between $X_{t-k:t}$ and $Y_{t+1}$ conditioned on the history of $Y$, i.e., $Y_{t-l:t}$. Clearly, different choices of $k$ and $l$ result in different outcomes when performing TE measurements on a dataset, therefore, it is necessary to find the optimal values of $k$ and $l$. For a 1$^{st}$-order Markov processes, the optimal values are obviously $k=l=1$. 
Furthermore, in order to precisely calculate transfer entropy from equation~\ref{eq:te}, the summation should be performed over all instances of variables $X_t$, $Y_t$, $Y_{t+1}$, and as result using only a subset of those instances may result in an estimate of the precise transfer entropy. This is another common source of inaccuracy in TE measurements of neural recordings. Here we were able to generate neural recording data for all possible sensory input patterns and included them in our dataset, yet still observe the described shortcomings in our results. This brings up another important point to notice, namely, even if we introduce every possible sensory pattern to the network, we do not necessarily observe every possible neural firing pattern in the network, and as a result, we do not necessarily span the entire set of variable states $(Y_{t+1}, Y_t, X_t)$.

Our results imply that transfer entropy has its own limitations in accurately estimating the information flow, and its accuracy may depend on the type of network or cognitive task it is applied to, as well as the type of data that is used to construct the measure. These findings highlight the importance of understanding the frequency and types of fundamental processes and relations in biological nervous systems. For example, one approach would be to examine causality detection methods such as transfer entropy in known systems, especially in known simple biological neural networks in order to shed light on strengths and deficiencies of current methods. Performing a causal analysis on brains in vivo will remain a daunting task for the foreseeable future, but advances in the evolution of digital cognitive systems may allow us a glimpse of those, and perhaps guide the development of other measures of information flow.
%%%%%%%%%%%%%%%%%%%%%%%%%%%%%%%%%%%%%%%%%%
%%%%%%%%%%%%%%%%%%%%%%%%%%%%%%%%%%%%%%%%%%
%\section{Conclusions}

%This section is not mandatory, but can be added to the manuscript if the discussion is unusually long or complex.

\vspace{6pt} 

%%%%%%%%%%%%%%%%%%%%%%%%%%%%%%%%%%%%%%%%%%
%\authorcontributions{``conceptualization, A.T. and C.A.; methodology, A.T. and C.A.; data production and analysis, A.T.; writing--original draft preparation, A.T.; writing--review and editing, A.T. and C.A.; funding acquisition, C. A.''}

%%%%%%%%%%%%%%%%%%%%%%%%%%%%%%%%%%%%%%%%%%
%\funding{This research was funded by the BEACON Center for the Study of Evolution in Action and the National Science Foundation under Cooperative Agreement No. DBI-0939454 at Michigan State University.}

%%%%%%%%%%%%%%%%%%%%%%%%%%%%%%%%%%%%%%%%%%
\section*{Acknowledgements}
We acknowledge computational resources provided by the Institute for Cyber-Enabled Research (iCER) at Michigan State University.

%%%%%%%%%%%%%%%%%%%%%%%%%%%%%%%%%%%%%%%%%%
%\conflictsofinterest{The authors declare no conflict of interest.The funders had no role in the design of the study; in the collection, analyses, or interpretation of data; in the writing of the manuscript, or in the decision to publish the results'.} 
%%%%%%%%%%%%%%%%%%%%%%%%%%%%%%%%%%%%%%%%%%
%% optional
\section*{Abbreviations}

The following abbreviations are used in this manuscript:\\

\noindent 
\begin{tabular}{@{}ll}
MB & Markov Brains\\
GA & Genetic Algorithm\\
TE & Transfer Entropy\\
PD & Preferred Direction\\
ND & Null Direction\\
MD & Motion Detection\\
SL & Sound Localization\\
XOR & Exclusive OR\\
XNOR& Exclusive NOR\\
SD  & standard deviation\\
SE  & standard error of mean\\
\end{tabular}

%%%%%%%%%%%%%%%%%%%%%%%%%%%%%%%%%%%%%%%%%%
%% optional
%\appendixtitles{no} %Leave argument "no" if all appendix headings stay EMPTY (then no dot is printed after "Appendix A"). If the appendix sections contain a heading then change the argument to "yes".
%\appendix
%\section{}
%\unskip

%%%%%%%%%%%%%%%%%%%%%%%%%%%%%%%%%%%%%%%%%%


%\externalbibliography{yes}
%\bibliography{bibliofile}
%\bliographystyle{unsrt}
%=====================================
% References, variant B: internal bibliography
%=====================================
%%%%%%%%%%%%%%%%%%%%%%%%%%%%%%%%%%%%%%%%%%
\end{document}